\documentclass{article} 
\usepackage{nips13submit_e,times}
\usepackage{hyperref}
\usepackage{url}
\usepackage{wrapfig}
\usepackage{graphicx}
\usepackage{amsmath}

\title{Sequence Modeling using Gated Recurrent Neural Networks}

\author{
Mohammad Pezeshki \\
Department of Computer Science and Operations Research\\
Universite de Montreal, Montreal QC H3C 3J7\\
\texttt{mohammadpz@gmail.com} \\
}

\nipsfinalcopy 

\begin{document}

\maketitle

\begin{abstract}
In this paper, we have used Recurrent Neural Networks to capture and model human motion data and generate motions by prediction of the next immediate data point at each time-step. Our RNN is armed with recently proposed Gated Recurrent Units which has shown promissing results in some sequence modeling problems such as Machine Translation and Speech Synthesis. We demonstrate that this model is able to capture long-term dependencies in data and generate realistic motions.
\end{abstract}

\section{Introduction}
Sequence modeling has been a challenging problem in Machine Learning that requires models which are able to capture temporal dependencies. One of the early models for sequence modeling was Hidden Markov Model (HMM) [1]. HMMs are able to capture data distribution using multinomial latent variables. In this model, each data point at time $t$ is conditioned on the hidden state at time $t$. And hidden state at time $t$ is conditioned on hidden state at time $t-1$. In HMMs both $P(x_t|s_t)$ and $P(s_t|s_{t-1})$ are same for all time-steps. A similar idea of \textit{parameter sharing} is used in Recurrent Neural Network (RNN) [2]. RNNs are an extention of feedforward neural networks which their weights are shared for every time-step in data. Consequently, we can apply RNNs to sequential input data. 

Theoretically, RNNs are capable of capturing sequences with arbitrary complexity. Unfortunately, as shown by Bengio et al. [3], there are some difficulties during training RNNs on sequences with long-term dependencies. Among lots of solutions for RNNs' training problems over past few decades, we use Gated Recurrent Units which is recently proposed by Cho et al. [4]. As it shown by Chung et al. [5], Gated Recurrent Unit performs much more better than conventional $Tanh$ units.

In the folowing sections, we are going to introduce the model, train it on the MIT motion database [6], and show that it is capable of capturing complexities of human body motions. Then we demonstrate that we are able to generate sequences of motions by predicting the next immediate data point given all previous data points.

\section{Recurrent Neural Network}
\label{gen_inst}

Simple Recurrent Neural Network which has been shown to be able to implement a Turing Machine [7] is an extension of feedforward neural networks. The idea in RNNs is that they share parameters for different time-steps. This idea which is called \textit{parameter sharing} enables RNNs to be used for sequential data.

RNNs have memory and can memorized input values for some period of time. More formally, if the input sequence is $\textbf{x} = \{x_1, x_2, ..., x_N\}$ then each hidden state is function of current input and previous hidden state.
\begin{align*}
h_{t} = F_{\theta}(h_{t-1}, x_t)
\end{align*}
which $F_{\theta}$ is a linear regression followed by a non-linearity.
\begin{align*}
h_{t} = \mathcal{H}(W_{hh}h_{t-1}, W_{xh}x_t)
\end{align*}
where $\mathcal{H}$ is a non-linear function which in Vanilla RNN is conventional $Tanh$. It can be easily shown using the above equation that each hidden state $h_t$ is a function of all previous inputs.
\begin{align*}
h_{t} = G_{\theta}(x_1, x_2, ..., x_t)
\end{align*}
where $G_{\theta}$ is a very complicated and non-linear function which summerizes all previous inputs in $h_t$. A trained $G_{\theta}$ puts more emphasis on some aspects of some of the previous inputs trying to minimize overal cost of the network.

Finally, in the case of real-valued outputs, output in time-step $t$ can be computed as follows
\begin{align*}
y_{t} = W_{hy}h_t
\end{align*}
Note that bias vectors are omitted to keep the notation simple. A graphical illustration of RNNs is shown in figure 1.
\begin{figure}[h]
\centering
\includegraphics[width=0.6\textwidth]{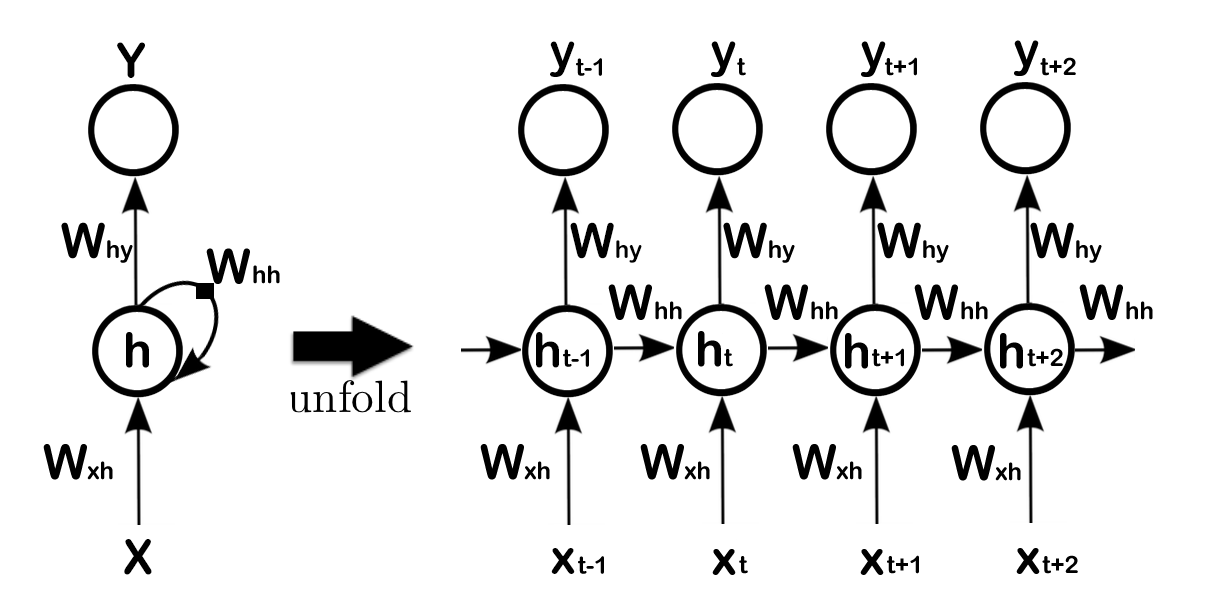}
\caption{\label{fig:frog}Left: A Recurrent Neural Network with recurrent connection from hidden units to themselves. Right: A same network but unfolded in time. Note that weight matrices are same for every time-step.}
\end{figure}

\subsection{Generative Recurrent Neural Network}
We can use a Recurrent Neural Network as a generative model in a way that the output of the network in time-step $t-1$ defines a probability distribution over the next input at time-step $t$. According to chain rule, we can write the joint probability distribution over the input sequence as follows.
\begin{align*}
P(x_1, x_2, ..., x_N) = P(x_1)P(x_2|x_1)...P(x_T|x_1, ..., X_{T-1})
\end{align*}
Now we can model each of these conditional probability distributions as a function of hidden states.

\begin{align*}
P(x_t|x_1, ..., x_{t-1}) = f(h_t)
\end{align*}
Obviously, since $h_t$ is a fixed length vector and $\{x_1, ..., x_{t-1}\}$ is a variable length sequence, it can be considered as a lossy compression. During learning process, the network should learn to keep important information (according to the cost function) and throw away useless information. Thus, in practice network just look at some time-steps back until $x_{t-k}$. The architecture of a Generative Recurrent Neural Network is shown in figure 2. 
\begin{figure}[h]
\centering
\includegraphics[width=0.4\textwidth]{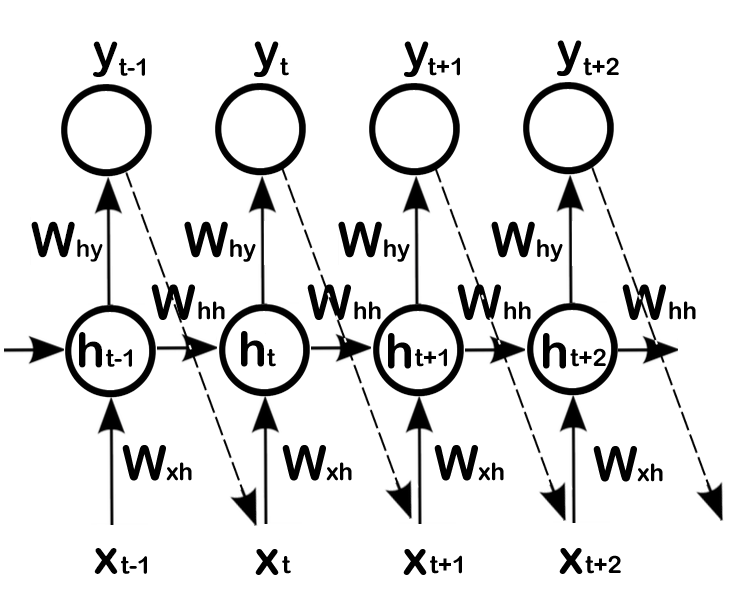}
\caption{\label{fig:frog}An unfolded Generative Recurrent Neural Networks which output at time-step $t-1$ defines a conditional probability distribution over the next input.  Dashed-lines are during generating phase.}
\end{figure}

Unfortunately, as shown by Bengio et al. [3], there are some optimization issues when we try to train such models with long-term dependency in data. The problem is that when an error occurs, as we back-propagate it through time to update the parameters, the gradient may decay exponentially to zero (Gradient Vanishing) or get exponentially large. For the problem of huge gradients, an ad-hoc solution is to restrict the gradient not to go over a threshold. This technique is known as \textit{gradient clipping}. But the solution for \textit{Gradient Vanishing} is not trivial. Over past few decades, several methods were proposed [e.g. 8, 9, 10] to tackle this problem. Although the problem still remains, gating methods have shown promissing results in comparison with Vanilla RNN in different task such as Speech Recognition [11], Machine Translation [12], and Image Caption Generation [13]. One of the models which exploits a gating mechanism is Gated Recurrent Unit [4].

\subsection{Gated Recurrent Unit}
Gated Recurrent Unit (GRU) is different from simple RNN in a sense that in GRU, each hidden unit has two gates. These gates are called update and reset gates which control the flow of information inside each hidden unit. Each hidden state at time-step $t$ is computed as follows,
\begin{align*}
h_{t} = (1-z_t)\circ h_{t-1} + z_{t}\circ \widetilde{h}_t
\end{align*}
where $\circ$ is an element wise product, $z_t$ is update gate, and $\widetilde{h}_t$ is the candidate activation.
\begin{align*}
\widetilde{h}_t = \tanh (W_{xh}x_t + W_{hh}(r_t\circ h_{t-1}))
\end{align*}
where $r_t$ is the reset gate. Both update and reset gates are computed using a sigmoid function:
\begin{align*}
z_t=\sigma(W_{xz}x_t+W_{hz}h_{t-1})\\
r_t=\sigma(W_{xr}x_t+W_{hr}h_{t-1})
\end{align*}
where $W$s are weight matrices for both gates. A Gated Recurrent Unit is shown in figure 3.
\begin{figure}[h]
\centering
\includegraphics[width=0.4\textwidth]{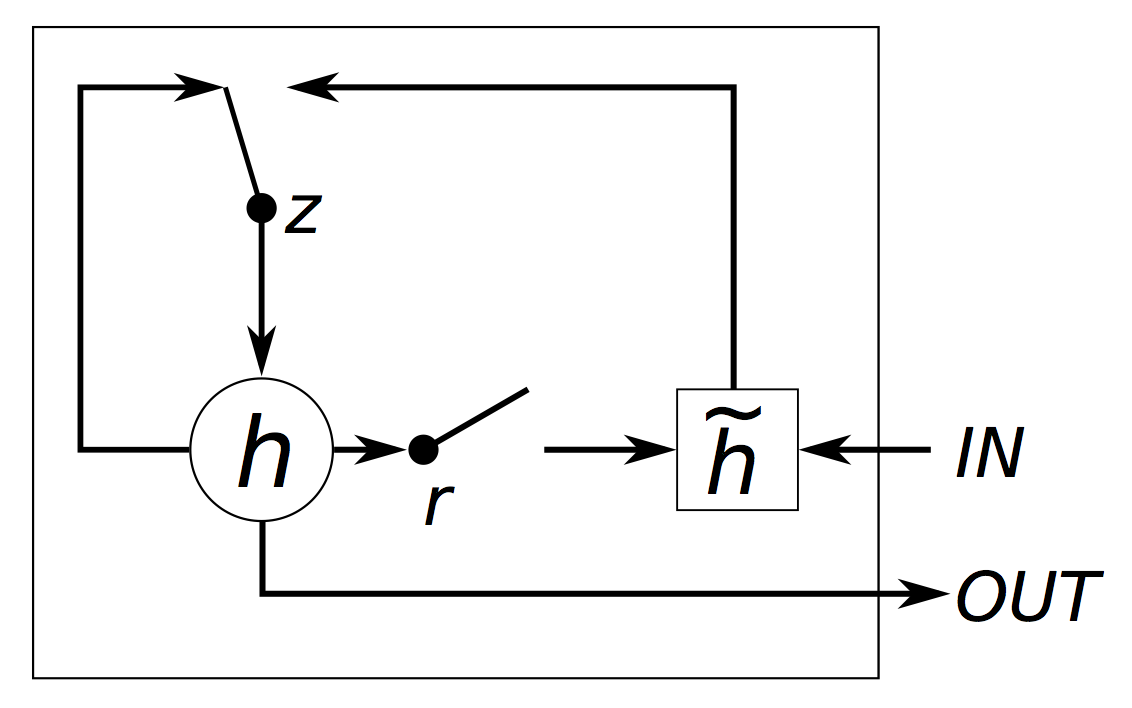}
\caption{\label{fig:frog}A Gated Recurrent Unit which $r$ and $z$ are the reset and update gates, and $h$ and $\widetilde{h}$ are the activation and the candidate activation. [5]}
\end{figure}

\section{Experimental results}
In this section we describe motion dataset and results for modeling and generating human motions.

\subsection{Ability to capture long-term dependency}
Before training model on motion data let's first compare GRU with conventional $Tanh$. As we discussed in section 2, due to optimization problems, simple RNNs are not able to capture long-term dependency in data (Gradient Vanishing problem). Thus, instead of using $Tanh$ activation function, we use GRU. Here we try to show that GRU performs much more better. The task is to read a sequence of random numbers, memorize them for some periods of time, and then emit a function which is sum over input value. We generated 100 different sequences, each containing 20 rows (time-steps) and 2 columns (attributes of each datapoint). We trained the models such that output at time $t$ ($y_t$) is a function of previous input values.
\begin{align*}
y_t = x_{t-3}[0] + x_(t-5)[1]
\end{align*}
Hence, we expect models to memorize 5 time-steps back and learn when to use which dimensions of the previous inputs. For both models, input is a vector of size 2, output is a scaler value, and a single hidden layer has 7 units. We allowed both networks to overfit on training data. It is shown in figure 4 that the model with GRU is able to perform very well while simple $Tanh$ cannot capture.
\begin{figure}[h]
\centering
\includegraphics[width=0.9\textwidth]{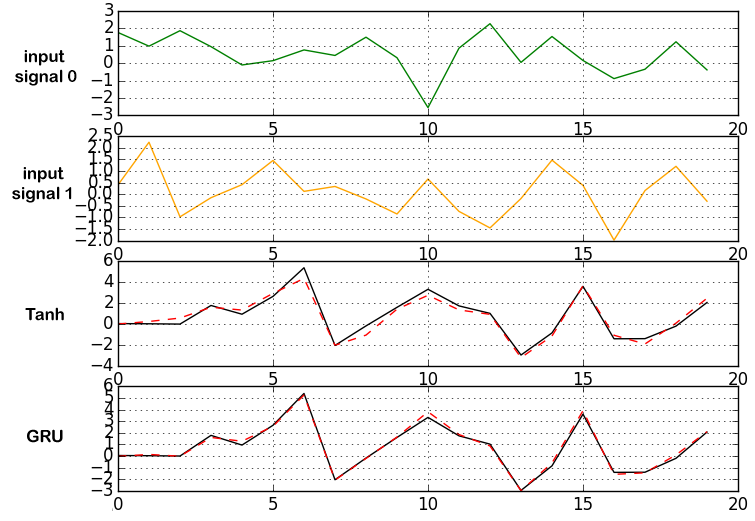}
\caption{\label{fig:frog}First two graphs show two input signals of the model for 20 time-steps. Third and fourth graphs are associated with $Tanh$ and GRU respectively. The solid line is real target and the dashed-line is the model output. As it is clear, the fourth graph performs much more better and is able to model target signal very well.}
\end{figure}

\subsection{Dataset}
Among some Motion Capture (MOCAP) datasets, we used simple walking motion from MIT Motion dataset [6]. The dataset is generated by filming a man wearing a cloth with 17 small lights which determine position of body joints. Each data point in our dataset consists of information about global orientation and displacement. To be able to generate more realistic motions, we used same preprocessing as used by Taylor et al. [14]. Our final dataset contains 375 rows where each row contains 49 ground-invarient, zero mean, and unit variance features of body joints during walking. We also used Neil Lawrence’s motion capture toolbox to visualize data in 3D spcae. Samples of data are shown in figure 5.
\begin{figure}[h]
\centering
\includegraphics[width=0.7\textwidth]{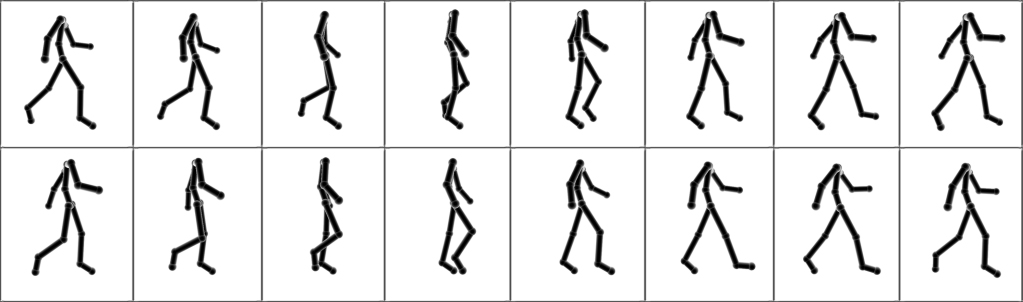}
\caption{\label{fig:frog}A sequence of frames of walking from MIT motion dataset. Each frame has 17 light points on the body. [6]}
\end{figure}

\subsection{Motion generation}
We trained our GRU Recurrent Neural Network which has 49 input units and 120 hidden units in a single hidden layer. Then, we use it in a generative fashion which each output at time $t$ is fed to the model as $x_{t+1}$. To initialize the model, we first feed the model with 50 frames of the training data and then let the model to generate arbitrary length sequence. Regeneration quality is good enough in a way that it cannot be distinguished from real trining data by the naked eye. In figure 6 average over all 49 features is ploted for better visualization. The initialization and generation phases are shown in figure 7.
\begin{figure}[h]
\centering
\includegraphics[width=0.7\textwidth]{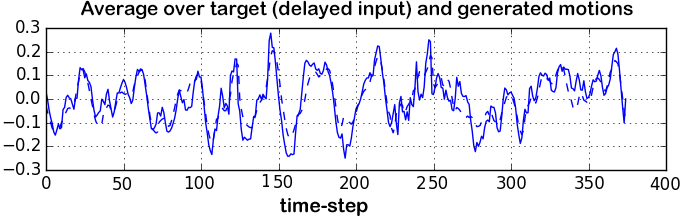}
\caption{\label{fig:frog}Average (for visualization) of all 49 features is shown in solid line for real target and in dashed-line for generated data.}
\end{figure}
\begin{figure}[h]
\centering
\includegraphics[width=0.7\textwidth]{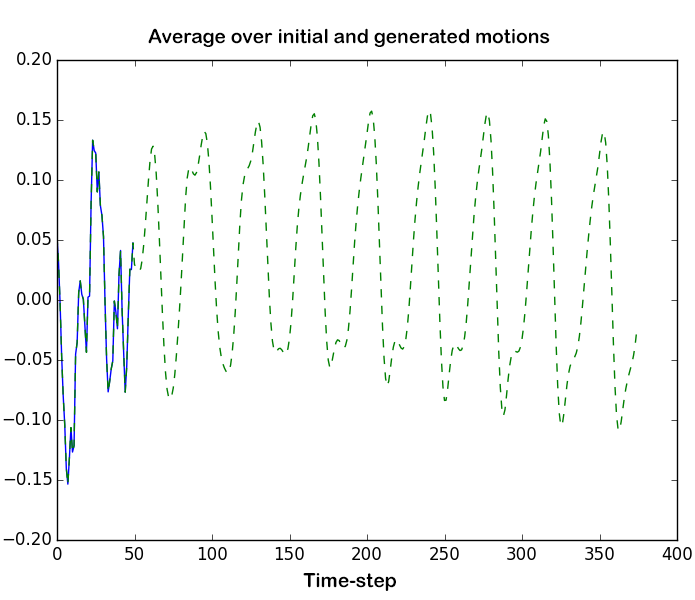}
\caption{\label{fig:frog}This graph shows the initialization using real data for first 50 time-steps in solid blue line and the rest is generated sequence in green dashed-line.}
\end{figure}

\section{Conclusion}
In this paper we have demonstrated that Gated Recurrent Unit helps optimization problems of Recurrent Neural Network when there is long-term dependency in data. We did our experiments discriminatively using a toy example dataset and generatively using MIT motion dataset and showed that GRU performs much better than simple Recurrent Neural Networks with conventional $Tanh$ activation function in both tasks of memorizing and generating.

\clearpage
\subsubsection*{References}

\small{
[1] Rabiner, Lawrence, and Biing-Hwang Juang. "An introduction to hidden Markov models." ASSP Magazine, IEEE 3.1 (1986): 4-16.

[2] Rumelhart, D. E., G. E. Hinton, and R. J. Williams. "Learning Internal Representations by Error Propagation, Parallel Distributed Processing, Explorations in the Microstructure of Cognition, ed. DE Rumelhart and J. McClelland. Vol. 1. 1986." (1986).

[3] Bengio, Yoshua, Patrice Simard, and Paolo Frasconi. "Learning long-term dependencies with gradient descent is difficult." Neural Networks, IEEE Transactions on 5.2 (1994): 157-166.

[4] 2012. URL http://icml.cc/discuss/2012/590.html.
K. Cho, B. van Merrienboer, D. Bahdanau, and Y. Bengio. On the properties of neural machine translation: Encoder-decoder approaches. arXiv preprint arXiv:1409.1259, 2014.

[5] Chung, Junyoung, et al. "Empirical Evaluation of Gated Recurrent Neural Networks on Sequence Modeling." arXiv preprint arXiv:1412.3555 (2014).

[6] Hsu, Eugene, Kari Pulli, and Jovan Popović. "Style translation for human motion." ACM Transactions on Graphics (TOG) 24.3 (2005): 1082-1089.

[7] Siegelmann, Hava T., and Eduardo D. Sontag. "Turing computability with neural nets." Applied Mathematics Letters 4.6 (1991): 77-80.

[8] Martens, James, and Ilya Sutskever. "Learning recurrent neural networks with hessian-free optimization." Proceedings of the 28th International Conference on Machine Learning (ICML-11). 2011.

[9] Pascanu, Razvan, Tomas Mikolov, and Yoshua Bengio. "On the difficulty of training recurrent neural networks." arXiv preprint arXiv:1211.5063 (2012).

[10] Bengio, Yoshua, Nicolas Boulanger-Lewandowski, and Razvan Pascanu. "Advances in optimizing recurrent networks." Acoustics, Speech and Signal Processing (ICASSP), 2013 IEEE International Conference on. IEEE, 2013.

[11] Graves, Alex, and Navdeep Jaitly. "Towards end-to-end speech recognition with recurrent neural networks." Proceedings of the 31st International Conference on Machine Learning (ICML-14). 2014.

[12] Bahdanau, Dzmitry, Kyunghyun Cho, and Yoshua Bengio. "Neural machine translation by jointly learning to align and translate." arXiv preprint arXiv:1409.0473 (2014).

[13] Vinyals, Oriol, et al. "Show and Tell: A Neural Image Caption Generator." arXiv preprint arXiv:1411.4555 (2014).

[14] Taylor, Graham W., Geoffrey E. Hinton, and Sam T. Roweis. "Modeling human motion using binary latent variables." Advances in neural information processing systems. 2006.

}

\end{document}